\documentclass[10pt,twocolumn,letterpaper]{article}

\usepackage{cvpr}
\usepackage{times}
\usepackage{epsfig}
\usepackage{graphicx}
\usepackage{amsmath}
\usepackage{amssymb}
\usepackage{algorithm, algorithmic}

\usepackage{stfloats}

\newlength\myindent
\usepackage[pagebackref=true,breaklinks=true,letterpaper=true,colorlinks,bookmarks=false]{hyperref}

\cvprfinalcopy 

\def\cvprPaperID{1122} 

\ifcvprfinal\fi
\begin{document}

\title{Structure Inference Machines: Recurrent Neural Networks for Analyzing Relations in Group Activity Recognition}

\author{Zhiwei Deng\qquad Arash Vahdat\qquad Hexiang Hu\qquad  Greg Mori\\School of Computer Science, Simon Fraser University, Canada\\
{\tt\small \{zhiweid, avahdat, hexiangh\}@sfu.ca, mori@cs.sfu.ca}
}

\date{September 2015}

\maketitle

\begin{abstract}

Rich semantic relations are important in a variety of visual recognition problems.  As a concrete example, group activity recognition involves the interactions and relative spatial relations of a set of people in a scene.
State of the art recognition methods center on deep learning approaches for training highly effective, complex classifiers for interpreting images.  However, bridging the relatively low-level concepts output by these methods to interpret higher-level compositional scenes remains a challenge.  Graphical models are a standard tool for this task.
In this paper, we propose a method to integrate graphical models and deep neural networks into a joint framework. Instead of using a traditional inference method, we use a sequential inference modeled by a recurrent neural network. Beyond this, the appropriate structure for inference can be learned by imposing gates on edges between nodes.  Empirical results on group activity recognition demonstrate the potential of this model to handle highly structured learning tasks.
   
\end{abstract}

\begin{figure}[htb]
\begin{center}
  \includegraphics[width=1.0\linewidth]{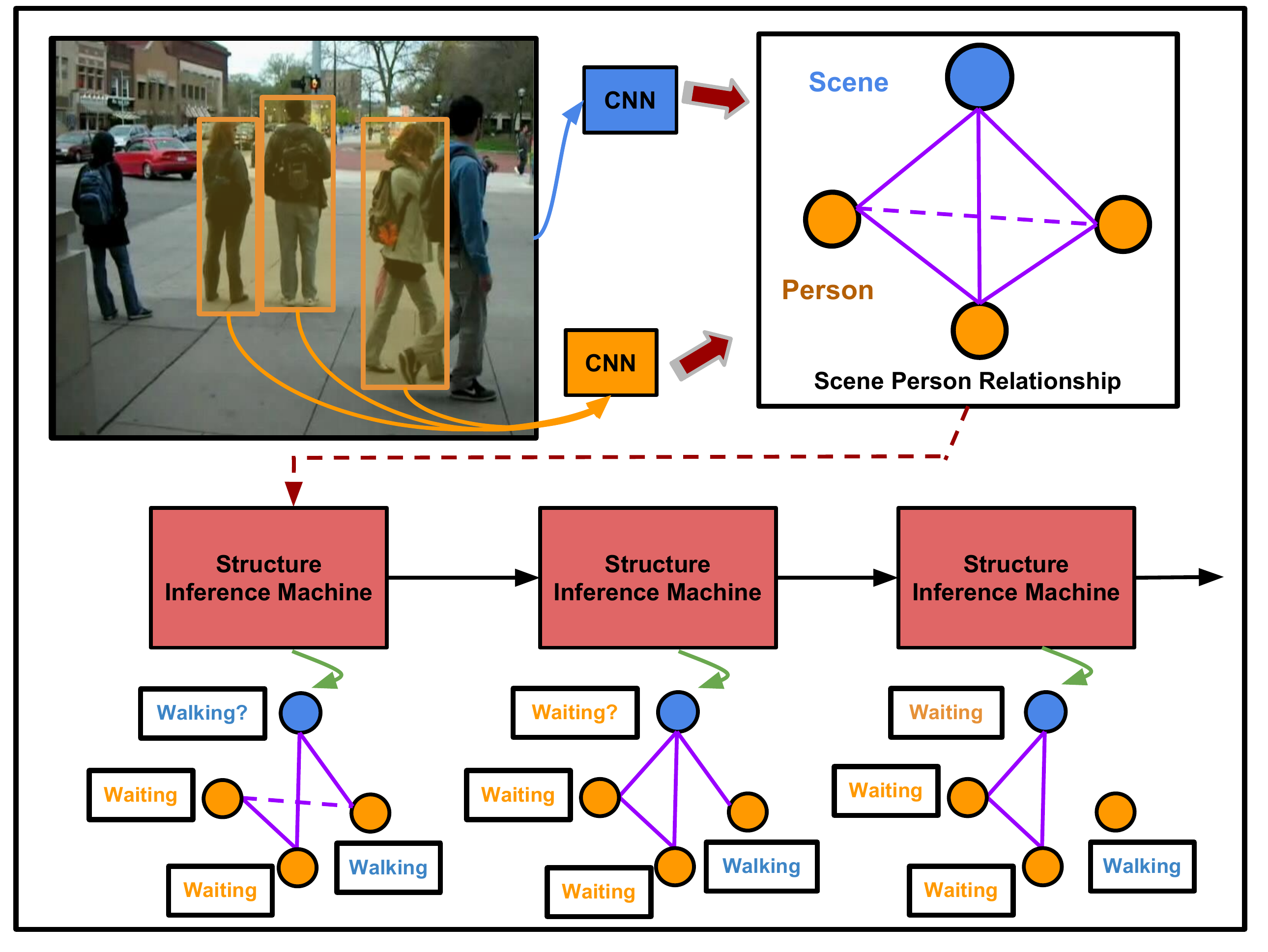}
\end{center}
\caption{Structure learning in a deep network.  Analyzing group activity requires reasoning about relations between the actions of individual people.  Our structure inference machine iteratively reasons about which people in a scene are interacting and which are involved in a group activity.}
\label{fig:pull-fig}
\end{figure}
\vspace{-3mm}

\section{Introduction}


Relations between image entities are an important facet of higher-level visual understanding.  Building relationships, such as the spatial distance between people in a scene, their relative motions, or concurrent actions can be used to drive recognition of higher-level activities.  Models for interpreting such scenes require the need to accurately interpret image cues, determine relevant relations between entities, and infer the properties of these relations.

In this paper we present a general-purpose method for this task, illustrated in Fig.~\ref{fig:pull-fig}.  The method builds upon deep networks for image analysis, endowing these networks with the ability to reason over structures and relationships.  This is accomplished by building higher-level recurrent networks that
equip the model with the ability to perform inference over lower-level network outputs, including learning structures that are effective for higher-level tasks.

We ground the work by developing specific models for group activity analysis.
Group activity analysis involves reasoning over individual people in a scene and considering their relations.  Multiple people in a scene could either be performing the same action at the same time, or have varied actions and interactions that compose a collective activity. Effective models need to jointly consider the rich relations between components of visual appearance. 

Standard approaches to this problem utilize graphical models to encode spatial relations and interactions. Recent work in this vein includes Choi et al.~\cite{ChoiCPS14}, who discover sub-groups of interacting people.  Lan et al.~\cite{LanSM12} proposed a hierarchical graphical model that considers the interactions on the social role level. Hajimirsadeghi and Mori \cite{HajimirsadeghiM15} proposed a gradient boosting training method. Amer et al.~\cite{AmerLT14} adopted a HiRF model to perform recognition and detection simultaneously. 
 

On another track, deep learning has proven successful in many computer vision applications, such as image classification, object recognition, and action recognition. On the image side, seminal work by Krizhevsky et al.~\cite{KrizhevskySH12} demonstrated the effectiveness of deep networks for object recognition; recent state of the art methods include GoogLeNet~\cite{Szegedy15}. On the video side, Simonyan and Zisserman~\cite{SimonyanZ14} proposed a two-stream convnet pipeline to apply deep learning to video analysis. Karpathy et al.~\cite{KarpathyTSLSF14} adopted various fusion techniques in convolutional neural networks to consider temporal information in video sequences. These methods have demonstrated the power of deep networks for classification tasks.

However, these models are trained to produce a flat classification output, categorizing an image/video according to the existence of a set of object/action labels.  For highly compositional tasks such as group activity recognition, models reasoning over structures can bring benefits, allowing the classification of higher-level concepts built from recognition over lower-level entities.

The main contribution of this paper centers on developing a model that bridges from low-level classifications to higher-level compositions.  We contribute an end-to-end trainable deep network that (1) classifies low-level image inputs according to their content, (2) refines these classifications by passing messages between outputs, (3) performs structure learning via gating functions that determine which outputs to connect, and (4) results in effective classification of high-level concepts.









\vspace{-2mm}
\section{Previous Work}
\vspace{-1mm}

This paper contributes a general-purpose deep learning inference machine and demonstrates its effectiveness for group activity recognition.  In this section we review relevant work on modeling structures in deep learning and specific models deployed for group activity recognition.


{\bf Deep Learning with Structures}:
Recently, there have been several interesting approaches to address the problem of combining graphical models and deep neural networks, primarily in the context of semantic image segmentation. 
Chen et al.~\cite{ChenPKMY15} proposed DeepLab, that feeds coarse responses at the final layer of a deep neural network to the CRF model proposed by Kr{\"a}henb{\"u}hl and Koltun \cite{KrahenbuhlK11}. 
Zheng et al.~\cite{Zheng15} proposed a CRF-RNN that trains the same model in an end-to-end fashion by transforming the approximate inference method \cite{KrahenbuhlK11} into a Recurrent Neural Network (RNN).
Schwing and Urtasun \cite{schwing2015fully} proposed an iterative procedure for end-to-end training of the CRF model.  We expand on this line of work by relaxing the assumptions that the underlying graphical model i) is fully connected and/or ii) has Gaussian kernels in the pairwise potential functions. Similarly, Chen et al.~\cite{chen2014learning} extended to general MRFs using approximate inference.  We use different inference techniques and enable structure learning.

Another line of work aims at modeling class relations in a graphical model that could be trained with a deep neural network. Deng et al.~\cite{Deng14large} proposed Hierarchy and Exclusion (HEX) graphs for modeling inclusion and exclusion relations between object classes and showed how these graphs could be used for computing HEX-based marginalized distributions of labels on top of a deep neural network.
Ding et al.~\cite{DingDMN15} extended HEX graphs to probabilistic HEX graphs modeled by Ising models and showed how standard Loopy Belief Propagation (LBP)
can be used in the inference. Both works rely on a special-purpose graphical model that represents particular relations between binary variables with a pre-designed structure. 

In the context of other structured problems, Bottou et al.~\cite{BottouBL97} proposed Graph Transformer Networks to jointly optimize subtasks. In this work, it was assumed that exact inference can be performed during a forward-backward pass. Ross et al.~\cite{RossMHB11} phrased structured prediction as a series of message passing steps.  Tompson et al.~\cite{tompson2014joint} proposed a feed-forward neural network that mimics a single iteration of the message passing algorithm for a markov random field for the task of human body pose recognition.  
Zhang et al.~\cite{zhang2015improving} incorporate structured prediction as a loss layer in a neural network.  Deng et al.~\cite{DengZCLMRM15} conduct message passing to do inference over a fixed structure for group activity recognition.  In contrast to these approaches, we infer structure via a gated network, allowing the model to determine the appropriate connections to use for inference.

{\bf Group Activity Recognition}:
Group activity recognition is typically modeled as a structured prediction problem that considers both individual actions and interactions with other people in a scene. Many previous work have used various forms of graphical models to address this problem: hierarchical graphical models \cite{amer2014hirf, LanSM12, LanWYRM12, RyooA11, choi_eccv12},
AND-OR graphs \cite{Amer2012cost, Gupta2009understanding} and  dynamic Bayesian networks \cite{zhu2013context} are among the popular models. 
Lan et al.~\cite{LanWYRM12} and Amer et al.~\cite{amer2014hirf} have shown the effectiveness of adaptive structures to the group activity recognition problem. Modeled by latent structure~\cite{LanWYRM12} or grouping nodes~\cite{amer2014hirf}, an adaptive structure
can adjust its structure to the most discriminative interactions in a scene. Khamis et al.~\cite{SamehCVPR12,SamehECCV12} utilize track-level and person-level features to determine group activity.
Shu et al.~\cite{ShuXRTZ15} reason about groups, roles, and events on top of noisy tracklets with a spatio-temporal AND-OR graph model.
Kwak et al.~\cite{KwakHH13} reason over temporal logic primitives in a quadratic programming formulation.
These models are trained in a sophisticated framework using shallow features and cannot easily be adopted in a deep learning framework.
In this work, we propose a general framework for integrating graphical models into a deep neural network that is capable of adapting their structure in an instance-based approach.

\begin{figure*}[t!]
\begin{center}
  \includegraphics[width=1.0\linewidth]{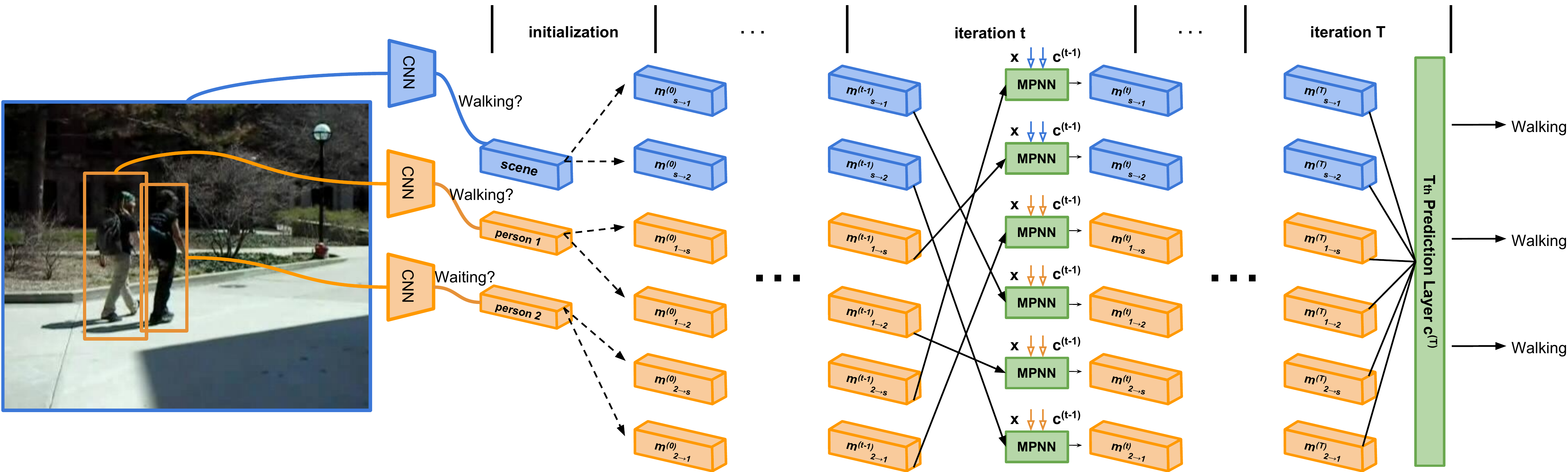}
\end{center}
\caption{The pipeline of inference in an RNN. We first use the unary scores to initialize the messages. In every iteration, new messages are computed using related message units, unary scores ($x$), and output predictions from the previous timestep ($c^{(t-1)}$). Note that for each timestep, a prediction layer outputs predictions (only illustrated in last layer), and in training receives loss as in a standard RNN.}
\label{fig:pipeline}
\vspace{-4mm}
\end{figure*}

\section{Structure Inference Machine}
\vspace{-1.5mm}

Group activity recognition requires reasoning about structures.  Interpreting an image of a scene of people involves determining what each individual person is doing and reasoning about their relations.  These tasks are both challenging due to ambiguity in image features and uncertainty in determining relations between people.  The ability to infer structures over the people in a scene is helpful for suppressing noise in the form of inaccurate human detections, mistaken low-level action recognition results, and spurious people not involved in a particular group activity.

Beyond group activity recognition, many other visual recognition tasks benefit from similar lines of reasoning.  Detecting and classifying individual component elements can be improved by considering structured relations among them.

The question we address in this paper is how to model such structured relations.  We take an approach building upon neural networks.  Deep learning-based methods have benefits in highly effective low-level action recognition, and we want to utilize this effectiveness within an end-to-end trainable model for higher-level reasoning.

Two components are used to cast this problem as a neural network formulation.
\begin{enumerate}
\item {\bf Recurrent neural networks for message passing}.  Consider an individual person within an image of a scene.  Ambiguity in inferring the action of the individual person is a fundamental problem.  As per standard arguments around context~\cite{TorralbaMFR03,HoiemEH06,Choi_VSWS_2009,LanWYRM12}, using the actions of other people in the scene can help to disambiguate the action of this individual.  We accomplish this by a recurrent neural network that aggregates cues about the actions of other people in a scene by repeatedly passing messages that refine estimates of an individual person's action.

\item {\bf Gating functions to learn structures}. Deciding who is interacting with whom in a group activity is an important part of inference.  Sub-groups of people can be engaged in different activities~\cite{ChoiCPS14}; individuals can be outliers compared to the group activity~\cite{LanWYRM12}. Reasoning over structures that determine connections between people in a scene can bring many benefits: which people are relevant to detecting the presence of an over-arching group activity, which people provide context for which others. Within a neural network structure, we accomplish this by introducing trainable gating functions that can turn on and off connections between individual people in the scene.
\end{enumerate}

Fig.~\ref{fig:pipeline} summarizes our structure inference machine.
The following sections present its details.  First, we present the use of recurrent neural networks as a tool for inference in a group activity model in Sec.~\ref{sec:bp_rnn}.  The use of gating functions to learn structure is presented in Sec.~\ref{sec:sl_rnn}.  By un-tying weights in these networks, we can relax assumptions regarding the message passing, leading to a general structure inference machine presented in Sec.~\ref{sec:dim}. 

\vspace{-1mm}
\section{Group Activity Recognition with an RNN}
\label{sec:bp_rnn}
\vspace{-1mm}

We build our model on top of a set of person detections in an image.  The model includes the actions for these individual people as well as the group activity for the whole image. Given a set of $M$ detected persons in a scene, a classifier (using a CNN) is used to provide visual classification $\{x_{i}\}_{i=1}^{M}$ of each person's action based on an image window cropped at the person detection. Each $x_{i}$ is a probability distribution over individual action for person $i$. In addition, a classifier that operates on the entire image can be used to directly estimate the group activity in the scene.  We denote by $x_s$ the group activity classification obtained from this whole-image classifier; $x_s$ is a distribution over possible scene-level group activities.

\begin{figure}
\begin{center}
  \includegraphics[width=0.9\linewidth]{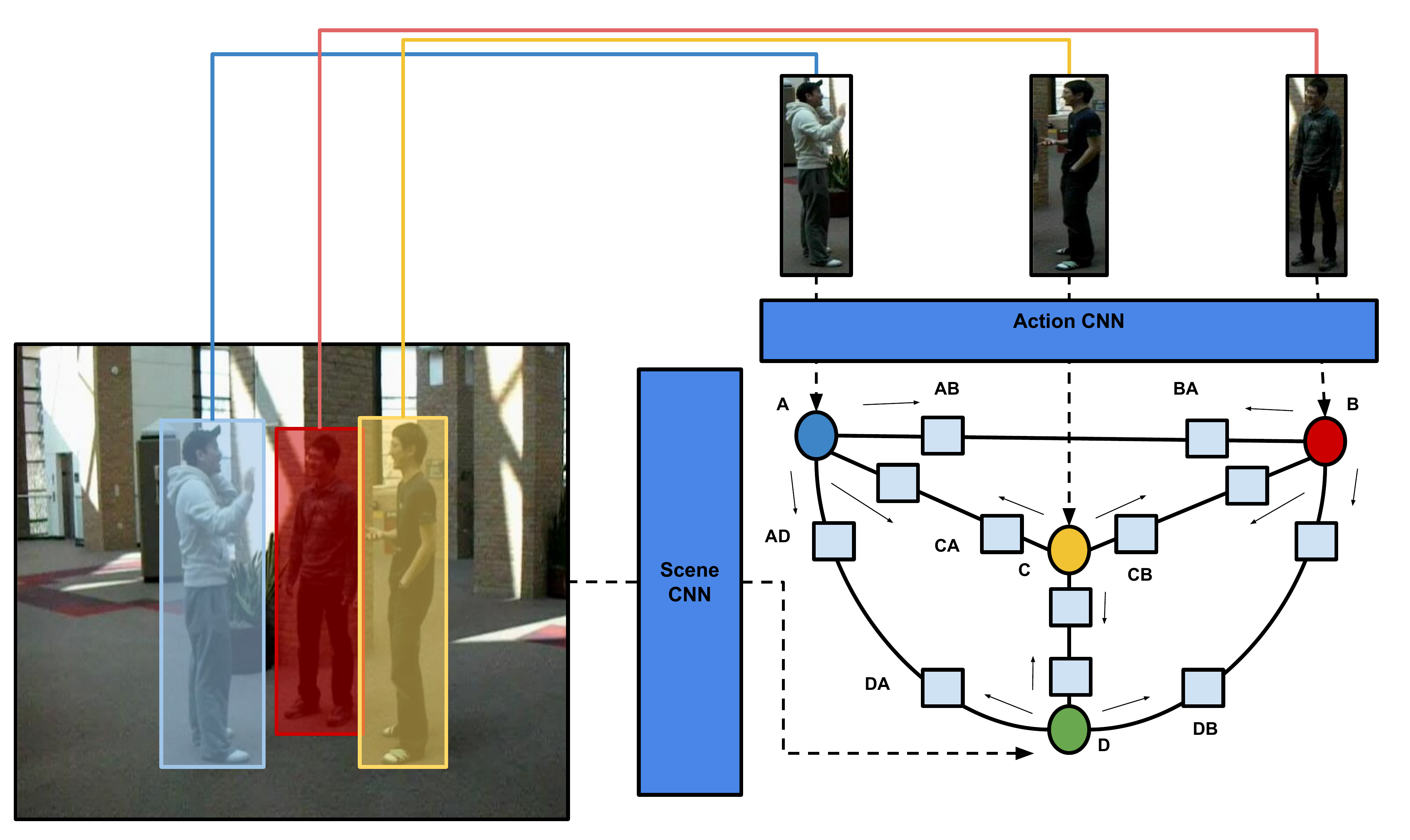}
\end{center}
\vspace{-2mm}
\caption{A group activity scenario represented as a graphical model.  Estimates of individual person actions and a group activity are refined via message passing. The squares are messages. These message units carry information from the source node and are propagated to the target node.}
\label{fig:crf}
\vspace{-5mm}
\end{figure}

A graphical model is built over these individual actions and the group activity, as shown in Fig.~\ref{fig:crf}. We use this graphical model to motivate our framework, and address structure learning in a graphical model with varied potential functions on top of a deep neural network.


\subsection{Recurrent Networks for Refining Action/Activity Classification}

On top of the individual classifications, we build a neural network structure which considers relations among entities to refine these classifications. This network structure is a recurrent neural network (RNN).  The RNN passes information amongst nodes representing individual person classifications and the scene classification in order to facilitate contextual refinement of classification decisions.


Given the graph structure (i.e.\ probabilistic graphical model in Fig.~\ref{fig:crf}), the RNN structure is built to model the connections between nodes. Distributions over values for each node $i$ can be determined by a combination of classification results $x_{i}$ of local observations, along with information passed along the graph.  We denote by $m$ information coming from related nodes along the graph.

In group activity recognition, this can model contextual relations between the actions of people in the scene. In every iteration of refinement, the values of the contextual information $m$ will be updated, while the inputs corresponding to visual observations of individual actions or scene label remain fixed.
 
We represent this iterative updating process as a recurrent neural network model.  In general, we take as input the local observations for all nodes $x = x_s \cup \{x_i: i=1,\ldots, M\}$.  We output a set of refined classification scores\footnote{These classification scores may in general be over different quantities from, or a subset of, the inputs.} $c^{(t)}$ at each timestep $t$.  These refined estimates are based on iteratively computed messages $m^{(t)}$ that pass information between the nodes:
\begin{eqnarray}
m^{(t)} &=& f(W_{mm}m^{(t-1)} + W_{xm}x + W_{cm}c^{(t-1)} + b_m) \hspace{0.6cm} \\
c^{(t)} &=& f(W_{mc}m^{(t)} + W_{xc}x + b_c)
\end{eqnarray}
In this and all subsequent equations, variables $W_{(\cdot)}$ and $b_{(\cdot)}$ refer to the neural network parameters which are to be learned. Here $f$ is an element-wise activation function for introducing non-linearity. At each time step, $x$, the scores from all image-based classifiers, is a static input into the recurrent neural network model. The ``hidden contextual information'' $m^{(t)}$ is updated by considering the previous contextual information $m^{(t-1)}$, the local observation $x$, and previous classification scores $c^{(t-1)}$. Each step of the RNN involves a single pass of aggregating information from contextual nodes within a graph to refine the scores of each node. Over a series of iterations, RNNs are used to allow finer refinements of these scores $c^{(t)}$.  



The formulation above is a general-purpose inference machine for refining estimates and producing classification outputs.  In the following section, we describe connections with graphical models and use this to motivate the specific choices of weight matrices and weight sharing we use for our group activity recognition model.
 


\subsection{Belief Propagation in an RNN}


Recall again the graphical model depicted in Fig.~\ref{fig:crf}.  This is a typical structure used in group activity recognition, with all people connected to each other and to the scene node corresponding to group activity~\cite{LanWYRM12,LanWYM10}.  For a probabilistic graphical model of this type, a standard inference algorithm is belief propagation -- rounds of passing messages between nodes would be conducted to obtain marginal distributions given observations.

We use the intuition from this probabilistic graphical model viewpoint to construct the specific form of weight matrices used in our structured inference machine.

In particular, in the graphical model formulation, we assume that all people are affecting each other and the scene. 
The message passing procedure starts by initializing all messages by unary energies from this graphical model. The unary energies are the inverse probability of an input image taking labels across action states, and are generated by a classifier (e.g.\ CNN). The initial messages are therefore independent probabilities without considering connections or smoothness between nodes. 

To model the connections between entities in a group activity, three types of weights are used: (1) the weights to map the relations from individual actions to scene-level group activity, (2) from scene-level group activity to individual actions, and (3) amongst individual actions of different people. This is analogous to typical potential functions which describe pairwise energies between actions and actions or actions and scenes (similar to the shared factors in ~\cite{DengZCLMRM15}). The pairwise connections provide a data-dependent smoothing term analyzing influence between entities and are crucial in understanding a highly structured problem.

\vspace{-5mm}

\subsubsection{Messages in an RNN}

\vspace{-1.5mm}

In the message passing recurrent neural network, each message unit is a vector composed of a set of neurons. The message passing is conducted between the message units. The configuration of connections in the RNN is determined by the graph structure/connections between nodes. Thus, each message unit's input is its neighbouring connected messages, the static unary inputs, and outputs from the previous time step. As the message units basically correspond to the distribution of the node~\cite{RossMHB11}, this process, roughly speaking, can be considered as classifying one entity by other related entities and the local observation of itself. We adopted a recurrent neural network structure which shares the weights of message computation in all time steps.


Consider a message $m_{i\rightarrow j}^{(t)}$. This message coming out of person $i$ on the edge connected to person $j$ corresponds to the distribution of $i$, and is classified by average pooled scores of neighbouring persons $(\sum_{k}m_{k \rightarrow i}^{(t-1)})/(|N_{i}|-1)$, static unary input $x_{i}$, and output $c_{i}^{(t-1)}$ from last time step. Denote the set of person nodes as $V^P$, the neighbouring person nodes of $i$ as $N_{i}$, $N_{i} \subseteq V^P$, the scene node as $s$ and the current time step as $t$. The mathematical form of $m_{i\rightarrow j}^{(t)}$ is:
\begin{equation}
\label{eqn:p2p}
\small
f\left(W_{mm^1} \left[x_{i},c_{i}^{(t-1)},\frac{\sum_{k}m_{k \rightarrow i}^{(t-1)}}{|N_{i}|-1}, m_{s\rightarrow i}^{(t-1)}\right]^T \right),\nonumber
\end{equation}
\begin{equation}
\small
i \in V^P, k \in N_{i} \backslash j
\end{equation}
where $W_{mm^1}$ is the concatenation of a set of weights: $W_{mm^1} = [W_{xm}, W_{cm}, W_{mm}^{(aa)}, W_{mm}^{(sa)}]$. Since message $m_{i\rightarrow j}$ corresponds to the distribution of node $i$, intuitively the weights $W_{mm}^{(aa)}$ and $W_{mm}^{(sa)}$ can be considered as classifying person $i$'s action based on other people's action or the scene label respectively. And the unary input $x_i$ and previous iteration's score $c_i^{(t-1)}$ are remapped via a linear transformation using $W_{xm}$ and $W_{cm}$. The function $f(\cdot)$ denotes the non-linear activation function, in our case simply a softmax function to normalize the message.


Likewise, if $i$ is a person node and $s$ is the scene node, then the message $m_{i\rightarrow s}^{(t)}$ is:
\begin{equation}
\label{eqn:p2s}
\small
f\left(W_{mm^2}\left[x_{i},c_{i}^{(t-1)},\frac{\sum_{k}m_{k\rightarrow i}^{(t-1)}}{|N_{i}|}\right]\right), k \in N_{i}
\end{equation}
where $W_{mm^2}$ equals $[W_{xm}, W_{cm}, W_{mm}^{(aa)}]$. Finally, if $s$ is the scene node, and $j$ is a person node, the message $m_{s\rightarrow j}^{(t)}$ is set as:
\begin{equation}
\label{eqn:s2p}
\small
f\left(W_{mm^3}\left[x_{s},c_{s}^{(t-1)},\frac{\sum_{k}m_{k\rightarrow s}^{(t-1)}}{|N_{s}|-1}\right]\right), k \in N_{s} \backslash p_j
\end{equation}
where $W_{mm^3}=[W_{xm},W_{cm},W_{mm}^{(as)}]$.



\vspace{-2mm}
\subsubsection{Output Prediction Layer} 
\vspace{-1mm}

The message units representing information all over the graph are formulated into the recurrent unit. However, to eventually classify a node in a graphical model, all messages around each node should be collected and used to perform prediction. In our model, we use a prediction layer to collect all messages around each node and infer a scene label for the group activity and action classification for each person. Through this layer, the losses on each time step are imposed on the message unit. More precisely, the prediction function for the scene level node is:
\begin{equation}
\label{pred:i}
\small
c_{s}^{(t)} = f\left(W_{hc^1}\left[x_{s}, \frac{\sum_{k}m_{k\rightarrow s}^{(t)}}{|N_{s}|}\right]\right),  p_k \in N_{s}
\end{equation}
where all notation is consistent with the previous message definitions. The activation function we used here is a softmax normalization.

Similarly, for performing action classification on a person-level node:
\begin{equation}
\label{pred:s}
\small
c_{i}^{(t)} = f\left(W_{hc^2}\left[x_{i}, \frac{\sum_{k}m_{k\rightarrow i}^{(t)}}{|N_{i}|}, m_{s\rightarrow i}^{(t)}\right]\right),  k \in N_{i}
\end{equation}

As shown above, the hidden contextual information is iteratively refined in a recurrent neural network through message passing. Similar to a standard belief propagation algorithm, the final prediction of each node is performed by collecting all related messages and is done in the prediction layer. The softmax loss imposed on the prediction results with standard mini-batch backpropagation training is used to train the model in an end-to-end framework.



\section{Structure Learning for Group Activity Inference}
\label{sec:sl_rnn}

\begin{algorithm}
\caption{Structure Inference Machine}
\label{alg}
\begin{algorithmic}
    \STATE \textbf{Inputs:} frame, detected person bounding boxes\\
    Pass image through CNN to get $x_s$\\
    Pass person bounding boxes through CNN to get $\{x_i\}_{i=1}^M$\\
    Initialize $m_{s\rightarrow i}^{(0)}$ by $x_s$; $ m_{i\rightarrow s}^{(0)},m_{i\rightarrow j}^{(0)}$ by  $\{x_i\}_{i=1}^M$
    \FOR {each iteration t}
    \FOR{edge $(i,j)$}
    \STATE Compute messages $m_{i\rightarrow j}^{(t)}$ and $m_{j\rightarrow i}^{(t)}$ by Eq.~\ref{eqn:p2p}-\ref{eqn:s2p}
    \ENDFOR
    \FOR{edge $(i,j)$}
    \STATE Compute gate value $g_{<i,j>}^{(t)}$ or $g_{<s,i>}^{(t)}$ by Eq.~\ref{gate:i2j}-\ref{gate:inj} \\
    Impose gates on $m_{i\rightarrow j}^{(t)}$ and $m_{j\rightarrow i}^{(t)}$ by Eq.~\ref{gate:gm}
    \ENDFOR
    \FOR{each node i}
    \STATE Compute node prediction $c_i^{(t)}$ by Eq.~\ref{pred:i},~\ref{pred:s}
    \ENDFOR
    \ENDFOR
    \STATE \textbf{Outputs: } Predicted scene label from timestep $T$, $c_s^{(T)}$
\end{algorithmic}
\end{algorithm}

For group activity, the structure of connections between people can greatly influence performance.  In general a fully connected graph, in which all people in a scene are connected to all others, and all people are related to the group activity, could model any type of relation.  However, including spurious edges relating irrelevant people introduces significant noise.  Instead focusing on relevant connections can lead to better models.

Beyond this, in a highly structured group activity, the connections between people or interactions between person and scene may vary according to different situation. Also, the interactions between people could be hard to explicitly capture. For example, in a crossing-the-street scene, both people crossing the street and people waiting for the lights to change are contributing to the group activity, while the people walking behind them may become irrelevant or even bring ambiguity. Hence, connectivity of the model should be able to adaptively change and adjust according to the particular input situation.

In the context of a recurrent neural network, gates are widely used as a tool for selecting information on the activation level. Both long short-term memory (LSTM) / gated recurrent units (GRUs) are proven to be successful on many tasks involving iteratively learning and gating information element-wise. In our model, we introduce the concept of ``instance level'' gates. An instance level gate is used to modify an edge of a graphical model which models the interactions between instances, such as a person to a person, or a person to a scene. Instead of using a vector gate to select information element-wise, we instead learn a scalar value gate function to enforce sparsity on the structure of a graphical model. 

Based on the previously introduced message passing RNN model, each node will receive information passed through an edge. Intuitively, the message passed could be noisy and may harm the understanding of actions of the instance. We propose to adopt an architecture similar to a LSTM gate, by taking multiple sources of information to selectively choose connectivity of nodes. To determine whether an edge is useful we compare the messages passed along this edge with other related messages. For example, to measure the gain by including the message $m_{A\rightarrow B}$, the message content of $m_{A\rightarrow B}$ is compared to other messages from other edges, the previous iteration's classification results, and the unary distribution on node $B$. The gain of including the edge $AB$ is the average value of gains for $m_{A\rightarrow B}$ and $m_{B\rightarrow A}$.  For two person nodes $i$ and $j$, the mathematical definition of the gating functions for message $m_{i\rightarrow j}$ is:
\begin{equation}
\label{gate:i2j}
\small
g_{i\rightarrow j}^{(t)} = \sigma\left(W_{hg^1}\left[x_{j},c_{j}^{(t-1)},m_{i\rightarrow j}^{(t)}, \frac{\sum_k m_{k\rightarrow j}(t)}{|N_{j}-1|} \right]\right), \nonumber
\end{equation}
\begin{equation}
\small
k \in N_{j} \backslash i
\end{equation}
where $W_{hg^1} = [W_{xg},W_{cg},W_{mm}^{(aa)},W_{mm}^{(aa)}]$. Here we reuse the weights $W_{mm}^{(aa)}$ which classifies a person's action by other people's action labels. The activation function $\sigma$ squeezes the gate value into $[0,1]$. In our model, we used a sigmoid activation function.

Similarly, for the scene node $s$ connecting to a person node $i$, the gate value for message $m_{s\rightarrow i}$ is calculated as:
\begin{equation}
\label{gate:s2i}
\small
g_{s\rightarrow i}^{(t)} = \sigma\left(W_{hg^2}\left[x_{i},c_{i}^{(t-1)},m_{s\rightarrow i}^{(t)}, \frac{\sum_k m_{k\rightarrow i}(t)}{|N_{i}|} \right]\right), k \in N_{i}
\end{equation}
where $W_{hg^2}$ is $[W_{xg},W_{cg},W_{mm}^{(sa)},W_{mm}^{(aa)}]$. And the gating function for $m_{i\rightarrow s}$ is:
\begin{equation}
\label{gate:i2s}
\small
g_{i\rightarrow s}^{(t)} = \sigma\left(W_{hg^3}\left[x_{s},c_{s}^{(t-1)},m_{i\rightarrow s}^{(t)}, \frac{\sum_k m_{k\rightarrow s}(t)}{|N_{s}|-1} \right]\right), k \in N_{s}
\end{equation}
where $W_{hg^3}$ equals $[W_{xg},W_{cg},W_{mm}^{(as)},W_{mm}^{(as)}]$. Then the gate values for an edge between a person and the scene $e_{<i, s>}$, and between two persons $e_{<i,j>}$, at timestep $t$, are calculated as:
\begin{equation}
\label{gate:ins}
\small
g_{<i, s>}^{(t)} = (g_{i\rightarrow s}^{(t)} + g_{s\rightarrow i}^{(t)})\slash 2
\end{equation}
\begin{equation}
\label{gate:inj}
\small
g_{<i, j>}^{(t)} = (g_{i\rightarrow j}^{(t)} + g_{j\rightarrow i}^{(t)})\slash 2
\end{equation}

After imposing the gates on relevant messages, the message units in the previous section are recalculated as:
\begin{equation}
\label{gate:gm}
\small
m\prime_{A\rightarrow B}^{(t)} = (g_{<A, B>}^{(t)} \odot m_{A\rightarrow B}^{(t)})
\end{equation}
Note that the above equation is a general form for gated messages. The symbol $\odot$ represents the product operation between a scalar and a vector. Nodes $A$ and $B$ could either be a person $i$ or a scene node $s$. 

Further more, to enforce the sparsity of connections between nodes and learn the most discriminative structures for each graph, we use the $L1$ regularization on the gate values. Given a particular training data sample (i.e.\ labeled frame) $d$, a graphical model is built on top of it as shown in above sections. Let $E^d$ denote the set of edges in the graph. The total loss on gates for the data sample $d$ is:
\begin{equation}
L^d = \lambda\sum_{e=1}^{|E^d|}(|g_e^d(\cdot)|)
\end{equation}
where $\lambda$ is the coefficient for this $L1$ regularization term to balance between the sparsity of the graph and prediction loss. $g_e^d(\cdot)$ is the gate value on edge $e$ of the graph for data sample $d$, where the gate could either be a scene-person or a person-person edge, or, in general the edge between two nodes. As the whole model is trained by the standard mini-batch method, the loss on each batch $B$ is $\sum_{d\in B}(L^d)$.


This structure selection is performed for each time step of message passing. An overall summary of the structure inference machine is presented in Alg.~\ref{alg}.

\subsection{Model Extension: Untying Weights as A Deep Inference Machine}
\label{sec:dim}

As shown previously, in the recurrent neural network framework, a graphical model described by various potential functions can be represented as weight-shared message predictors. However, by untying the weights of message computation for each step, the model could be further extended to a deep inference machine with structure gates to selectively pass information. A model with high non-linearity could be learned through this inference process.

In summary, this approach provides a general framework for both performing message passing of a RNN built from a graphical model and learning structures of a graphical model.  If the weights in the message passing steps are tied over iterations, this has direct analogy to inference in a graphical model.  If the assumption of tying of weights is relaxed, instead this process corresponds to a general deep inference machine with structure learning.


\section{Experiments}

We demonstrate our learning framework on group activity recognition. We provide results on three challenging datasets: (1) Collective Activity Dataset~\cite{Choi_VSWS_2009}; (2) Collective Activity Extended Dataset~\cite{Choi_CVPR_2011}; and (3) Nursing Home Dataset~\cite{DengZCLMRM15}.

The first two datasets are standard benchmarks widely used for group activity recognition. The Collective Activity Dataset contains 44 videos from 5 group activities (Crossing, Waiting, Queueing, Walking and Talking) and 6 individual actions (NA, Crossing, Waiting, Queueing, Walking and Talking). Collective Activity Extended omits the walking activity, due to ambiguities in its definition, and includes Jogging and Dancing categories. We follow the common protocol in \cite{LanWYRM12} for the Collective Activity Dataset. The scene label for a frame is defined by choosing the activity in which the most people participate. 

The Nursing Home Event Dataset contains human activities captured from fixed cameras in various rooms of a nursing home. It contains 80 videos showing 6 actions (walking, standing, bending, squating, sitting, falling) and two scenes (fall, non-fall). This dataset consists of many chanllenging frames with highly cluttered scenes and large intra-class variation within actions. We adopted the same protocol used in Deng et al.~\cite{DengZCLMRM15} for evaluation.

\textbf{Implementation details: } Our models are implemented using the Caffe library~\cite{jia2014caffe}. To acquire the action scores for each person image patch or the whole frame scene score, we fine-tuned the AlexNet architecture~\cite{KrizhevskySH12} pre-trained using the ImageNet data. We assume that an image has been pre-processed by a person detector to get person image patches. The message passing recurrent neural network is trained by adding a softmax loss on top of the output from each timestep. We also found it easier to train the gates by first fixing the weights of learned predictors and then learning structures on it. The number of neurons for the RNN per layer is $\sum_{AB\in\varepsilon_G}(|S_A|+|S_B|)$, where $\varepsilon_G$ is the set of edges of graphical model $G$, and $|S_A|$ and $|S_B|$ are the numbers of states of nodes $A$ and $B$ respectively.

\subsection{Collective Activity Dataset}
\vspace{-1mm}

We compare results of four different methods introduced in our paper with standard baselines. Table~\ref{sec:exp:cad} provides an ablation study examining the effects of different variants/components of our model.
There is a clear benefit by adopting structure gates to adaptively capture connections between nodes. Note that all the CRFs in our experiments are tuned on the validation set. Our model improved the accuracy of nodes over the whole graph: person-level action classification is improved by $\approx6\%$ after three steps of gated message passing.

\begin{table}
\begin{tabular}{|c|c|c|c|}
\hline
Iterations & 1 & 2 & 3\\
\hline
\hline
CRF + CNN & \multicolumn{3}{|c|}{74.18\%} \\
\hline
Struct. SVM + CNN & \multicolumn{3}{|c|}{73.87\%} \\
\hline
Tied Weights & 73.86\% & 74.02\% & 74.02\%\\
\hline
Untied Weights & 73.86\% &  74.33\% & 74.33\%\\
\hline
Gated Tied Weights & 80.12\% &  80.90\%  & \bf 81.22\% \\
\hline
Gated Untied Weights & 80.12\% &  81.06\% & \bf 81.22\%  \\
\hline
\end{tabular}
\caption{Results on Collective Activity Dataset.  Ablation study including variants of our model.}
\label{sec:exp:cad}
\vspace{-1mm}
\end{table}

Our model is compared to state of the art methods in Table \ref{sec:exp:cad_comp}.  The results are superior to other deep learning and structure learning models. The method of \cite{HajimirsadeghiMcvpr15} achieves better results, though uses a counting kernel (cardinality kernel) which directly mimics the majority-action scene label definition of Collective Activity Dataset.

\begin{table}
\begin{tabular}{|c|c|}
\hline
Method & Accuracy \\
\hline
Learning Latent Constituent \cite{AnticO14} & 75.1\% \\
\hline
Latent SVM with Optimized Graph \cite{LanWYRM12} & 79.7\% \\
\hline
Deep Struct. Model \cite{DengZCLMRM15} & 80.6\% \\
\hline
Unified Tracking And Recognition\cite{choi_eccv12} & 80.6\% \\
\hline
Cardinality Kernel \cite{HajimirsadeghiMcvpr15} & \bf 83.4\% \\
\hline
Our Model & 81.2\%  \\
\hline
\end{tabular}
\caption{Comparison with state-of-the-art methods on Collective Activity Dataset.}
\label{sec:exp:cad_comp}
\vspace{-5mm}
\end{table}

\vspace{-1mm}

\begin{figure*}[t]
\begin{center}
  \includegraphics[width=0.95\textwidth,scale = 0.5]{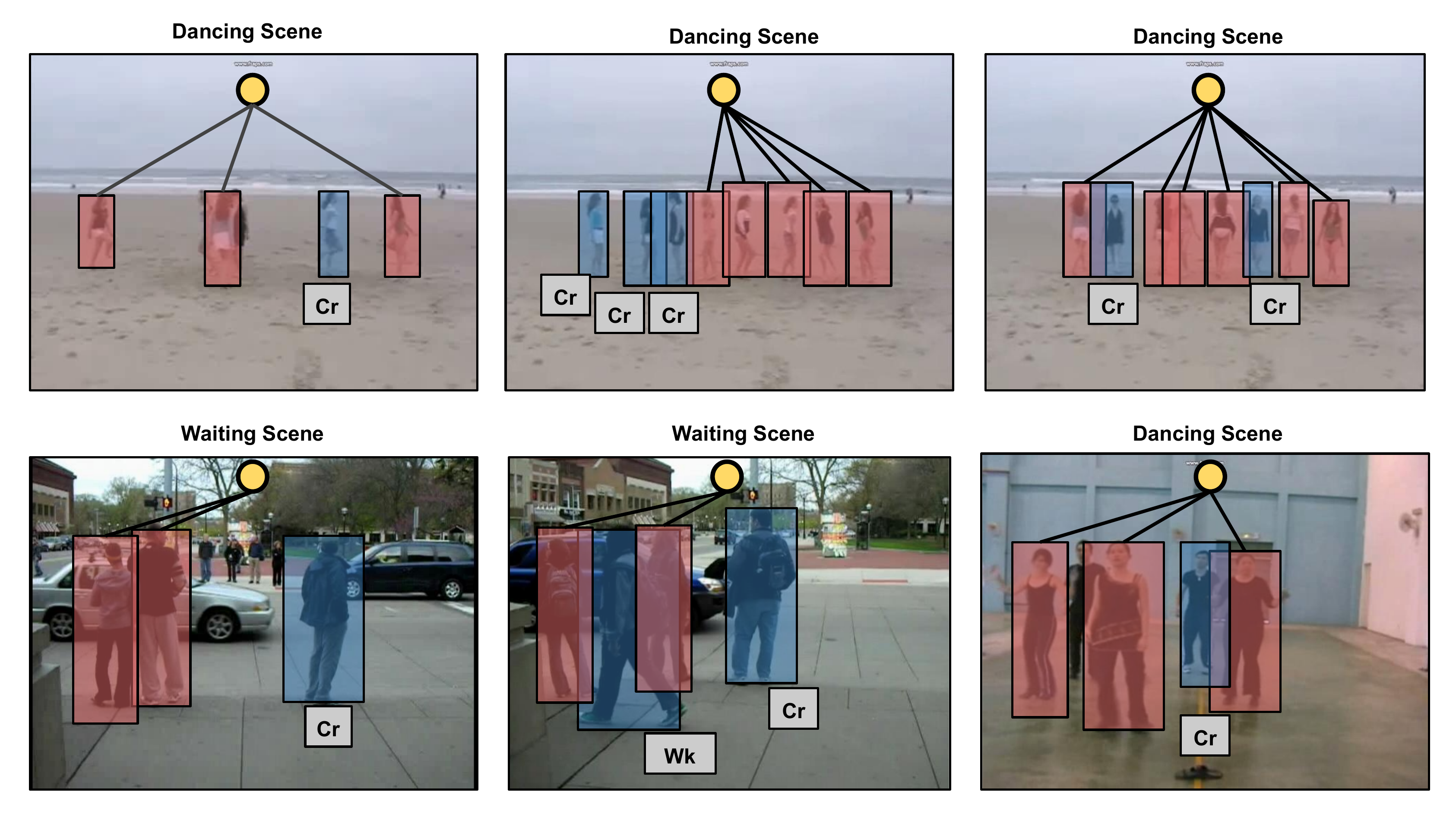}
\end{center}
\vspace{-5mm}
\caption{This figure shows visualizations of our experimental results. Note that these images are all misclassified by the fully connected graphical model. We show the scene gates learned in our model after 3 iterations of message passing and structure learning. For visualization, since the gate values are not strictly 0 or 1, we consider $<0.2$ as irrelevant/noisy connection versus $>0.7$ as useful connections. The red box has the same action class as the scene level node. Labels: ``Cr": Crossing, ``Wk": Walking.}
\label{fig:vis}
\vspace{-1mm}
\end{figure*}

\subsection{Collective Activity Extended Dataset}
We also experiment with the Collective Activity Extended Dataset. As noted in Choi et al.~\cite{Choi_CVPR_2011}, the walking action is ill-defined, hence we remove it, and include the  new actions Jogging and Dancing. 
The two previous works choose to adopt leave-one-out for testing. However, this is very computationally intensive for a  deep learning framework, and further makes hyper-parameter tuning a challenge. We choose to adopt a new train-test split with 2241 frames as training and 1106 as testing. 


Results are shown in  Table~\ref{sec:exp:cad_ext}. Note that on each person, action classification is improved by $\approx10\%$ via the structure inference process.
\begin{table}
\begin{tabular}{|c|c|c|c|}
\hline
Iterations & 1 & 2 & 3\\
\hline
\hline
CRF + CNN& \multicolumn{3}{|c|}{86.75\%} \\
\hline
Struct. SVM + CNN & \multicolumn{3}{|c|}{87.34\%} \\
\hline
Tied Weights & 84.45\% & 87.97\% & 87.97\% \\
\hline
Untied Weights & 84.45\% &  88.16\% & 88.16\% \\
\hline
Gated Tied Weights & 89.51\% &  90.14\%  & 90.14\% \\
\hline
Gated Untied Weights & 89.51\% &  90.14\% & \bf 90.23\% \\
\hline
\end{tabular}
\caption{Results on Collective Activity Extended Dataset.}
\label{sec:exp:cad_ext}
\vspace{-4mm}
\end{table}
\vspace{-0.5mm}

\subsection{Nursing Home Dataset}
On the Nursing Home Dataset, our person-level action classification accuracy also improved, by $\approx4\%$ after the second iteration. 
The accuracy is superior to baselines including Deng et al.~\cite{DengZCLMRM15}.
Note that in the Nursing Home Dataset, there is a smaller margin of improvement by adopting gating functions to learn structures. Because in each scene the irrelevant actions, such as sitting, can be identified as non-useful by simply a fully connected graphical model more easily than in the previous two datasets.

\begin{table}
\begin{tabular}{|c|c|c|c|}
\hline
Iterations & 1 & 2 & 3 \\
\hline
\hline
CRF + CNN & \multicolumn{3}{|c|}{83.64\%} \\
\hline
Struct. SVM + CNN & \multicolumn{3}{|c|}{82.08\%} \\
\hline
Deep Struct. Model \cite{DengZCLMRM15} & \multicolumn{3}{|c|}{84.7\%} \\
\hline
Tied Weights & 83.68\% & 84.91\% & 84.91\% \\
\hline
Untied Weights & 83.68\% &  84.94\% & 84.94\% \\
\hline
Gated Tied Weights & 84.46\% &  85.32\%  & 85.32\% \\
\hline
Gated Untied Weights & 84.46\% & \bf 85.50\% & \bf 85.50\% \\
\hline
\end{tabular}
\caption{Results on Nursing Home Dataset.}
\label{sec:exp:nursing}
\vspace{-5mm}
\end{table}

\vspace{-2mm}
\section{Conclusion}
\vspace{-1mm}

We presented a method for performing structure learning within a deep learning setting.  An inference algorithm for refining estimates of individual nodes and determining connections between nodes is implemented using a recurrent neural network with gating functions.  This approach was used to build a model for group activity recognition -- connections between individual people in a scene and their relation to the overarching scene-level activity label are learned.  This leads to improvements in accuracy over rounds of inference and structure learning via gating functions.

{\small
\bibliographystyle{ieee}
\bibliography{egbib}
}

\end{document}